\documentclass[runningheads]{llncs}

\usepackage[utf8]{inputenc}

\usepackage{graphics} 
\usepackage{epsfig} 
\usepackage{times} 
\usepackage{amsmath} 
\usepackage{amssymb}  
\usepackage{multirow}
\usepackage{lineno}
\usepackage{tikzpagenodes}
\usepackage{multirow}
\usepackage{bigdelim}
\usepackage{subfigure}
\usepackage{float}
\usepackage{physics}
\usepackage{hyperref}

\usepackage[dvips]{rotating} 
\usepackage{array}
\usepackage{multirow}
\usepackage{subfigure}
\usepackage{rotating}

\usepackage{textcomp}
\usepackage{epsfig}
\usepackage{changes}

\title{MixUp-MIL: Novel Data Augmentation for \\ Multiple Instance Learning and a Study on \\ Thyroid Cancer Diagnosis}

\titlerunning{MixUp-MIL: Novel Data Augmentation for Multiple Instance Learning}

\author{Michael Gadermayr$^1$ \and Lukas Koller$^1$ \and Maximilian Tschuchnig$^1$ \and Lea Maria Stangassinger$^2$ \and Christina Kreutzer$^3$ \and Sebastien Couillard-Despres$^3$ \and Gertie Janneke Oostingh$^2$ \and Anton Hittmair$^4$}

\authorrunning{M. Gadermayr et al.}

\institute{
$^1$ Salzburg University of Applied Sciences, Dept. of Information Technology and Digitalization\\
$^2$ Salzburg University of Applied Sciences, Dept. of Biomedical Sciences\\
$^3$ Spinal Cord Injury and Tissue Regeneration Center Salzburg, Research Institute of Experimental Neuroregeneration\\
$^4$ Kardinal Schwarzenberg Klinikum, Department of Pathology and Microbiology
}

\begin{document}

\maketitle

\begin{abstract}
	Multiple instance learning is a powerful approach for whole slide image-based diagnosis in the absence of pixel- or patch-level annotations.
	In spite of the huge size of whole slide images, the number of individual slides is often rather small, leading to a small number of labeled samples.
	To improve training, we propose and investigate novel data augmentation strategies for multiple instance learning based on the idea of linear and multilinear interpolation of feature vectors within and between individual whole slide images.
	Based on state-of-the-art multiple instance learning architectures and two thyroid cancer data sets, an exhaustive study was conducted considering a range of common data augmentation strategies.
	Whereas a strategy based on to the original MixUp approach showed decreases in accuracy, a novel multilinear intra-slide interpolation method led to consistent increases in accuracy.
\end{abstract}


\section{Motivation}\label{sec:introduction}
Whole slide imaging is capable of effectively digitizing specimen slides, showing both the microscopic detail and the larger context, without any significant manual effort.
Due to the enormous resolution of the whole slide images (WSIs), a classification based on straight-forward convolutional neural network architectures is not feasible.
%
Multiple instance learning~\cite{Ilse18a,Li21a,Wang2021,Zhang2022,Shao2021} (MIL) represents a methodology (with a high momentum indicated by a large number of recent publications) to deal with these huge images corresponding to single (global) labels.
In the MIL setting, WSIs correspond to labeled bags, whereas extracted patches correspond to unlabeled bag instances.
MIL approaches typically consist of a feature extraction stage, a MIL pooling stage and a following downstream classification.
State-of-the-art approaches mainly rely on convolutional neural network architectures for feature extraction, often in combination with attention~\cite{Li21b,Li21a} or self-attention~\cite{Rymarczyk2021}.
For training the feature extraction stage, classical supervised and self-supervised learning is employed~\cite{Li21b,Li21a}.
While the majority of methods rely on separate learning stages, also end-to-end approaches have been proposed~\cite{Chikontwe2020,Sharma2021}.
In spite of the large amount of data, the number of labeled samples in MIL (represented by the number of individual, globally labelled WSIs) is often small and/or imbalanced~\cite{Galdran2021}. 
General data augmentation strategies, such as rotations, flipping, stain augmentation and normalization and affine transformations, are applicable to increase the amount of data~\cite{Tellez2019}.
All of these methods are performed in the image domain.
Here, we consider feature-level data augmentation directly applied to the representation extracted using a convolutional neural network.
These methods can be easily combined with image-based augmentation and show the advantage of a high computational efficiency (since operations are efficient and pre-computed features can be used)~\cite{Li21a,Li21b}.
For example, Li et al.~\cite{Li21b} proposed an augmentation strategy based on sampling the patch-descriptors to generate several bags for an individual WSI.
In this paper, we focus on the interpolations of patch descriptors based on the idea of Zhang et al~\cite{Zhang2017a}, which is referred to as MixUp.
This method was originally proposed as data agnostic approach which also shows good results if applied to image data~\cite{Dabouei2021,Chen2022,Thulasidasan2019}. 
Variations were proposed, to be applied to latent representations~\cite{Verma2019} as well as to balance data sets~\cite{Galdran2021}.
Due to the structure of MIL training data, we identified several options to perform interpolation-based data augmentation.

The main contribution of this work is a set of novel data augmentation strategies for MIL, based on the interpolation of patch descriptors.
Inspired by the (linear) MixUp approach~\cite{Zhang2017a}, we investigated several ways to translate this idea to the MIL setting. 
Beyond linear interpolation, we also defined a more flexible and novel multilinear approach.
For evaluation, a large experimental study was conducted, including 2 histological data sets, 5 deep learning configurations for MIL, 3 common data augmentation strategies and 4 MixUp settings.
We investigated the classification of WSIs containing thyroid cancer tissues~\cite{Buddhavarapu2020,Gadermayr21a}.
To obtain an improved understanding of reasons behind the experimental results, we also investigate the feature distributions.

\section{Methods}
In this paper, we consider MIL approaches relying on separately trained feature extraction and classification stages~\cite{Li21a,Lerousseau21a,Rymarczyk2021}.
The proposed augmentation methods are applied to the patch descriptors obtained after the feature extraction stage.
This strategy is highly efficient during training since the features are only computed once (per patch) and for augmentation only simple arithmetic operations are applied to the (smaller) feature vectors.
Image-based data augmentation strategies (such as stain-augmentation, rotations or deformations) can be combined easily with the feature-based approaches but require individual feature extraction during training.
However, to avoid the curse of meta-parameters and thereby experiments these methods are not considered here.

In the original MixUp formulation of Zhang et al.~\cite{Zhang2017a}, synthetic samples $\boldsymbol{x'}$ are generated such that
$\boldsymbol{x'}  =\alpha \cdot \boldsymbol{x_i} + (1 - \alpha) \cdot \boldsymbol{x_j} \; ,$ 
where $\boldsymbol{x_i}$ and $\boldsymbol{x_j}$ are randomly sampled raw input feature vectors. Corresponding labels $\boldsymbol{y'}$ are generated such that
$\boldsymbol{y'} = \alpha \cdot \boldsymbol{y_i} + (1 - \alpha) \cdot \boldsymbol{y_j} \; ,$ 
where $\boldsymbol{y_i}$ and $\boldsymbol{y_j}$ are the corresponding one-hot label encodings. The weight $\alpha$ is drawn from a uniform distribution between $0$ and $1$.

A single input (corresponding to a WSI) of a MIL approach with a separate feature extraction stage~\cite{Li21a} can be expressed as a P-tupel $X = (\boldsymbol{x_1}, ..., \boldsymbol{x_P})$ with $\boldsymbol{x_i}$ being the feature vector of an individual patch and $P$ being the number of patches per WSI. 
The method proposed by Zhang et al. cannot directly be applied to these tupels. However, there are several options to adapt the basic idea to the changed setting.

\begin{figure}[tb] \center
	\includegraphics[width=0.99\linewidth]{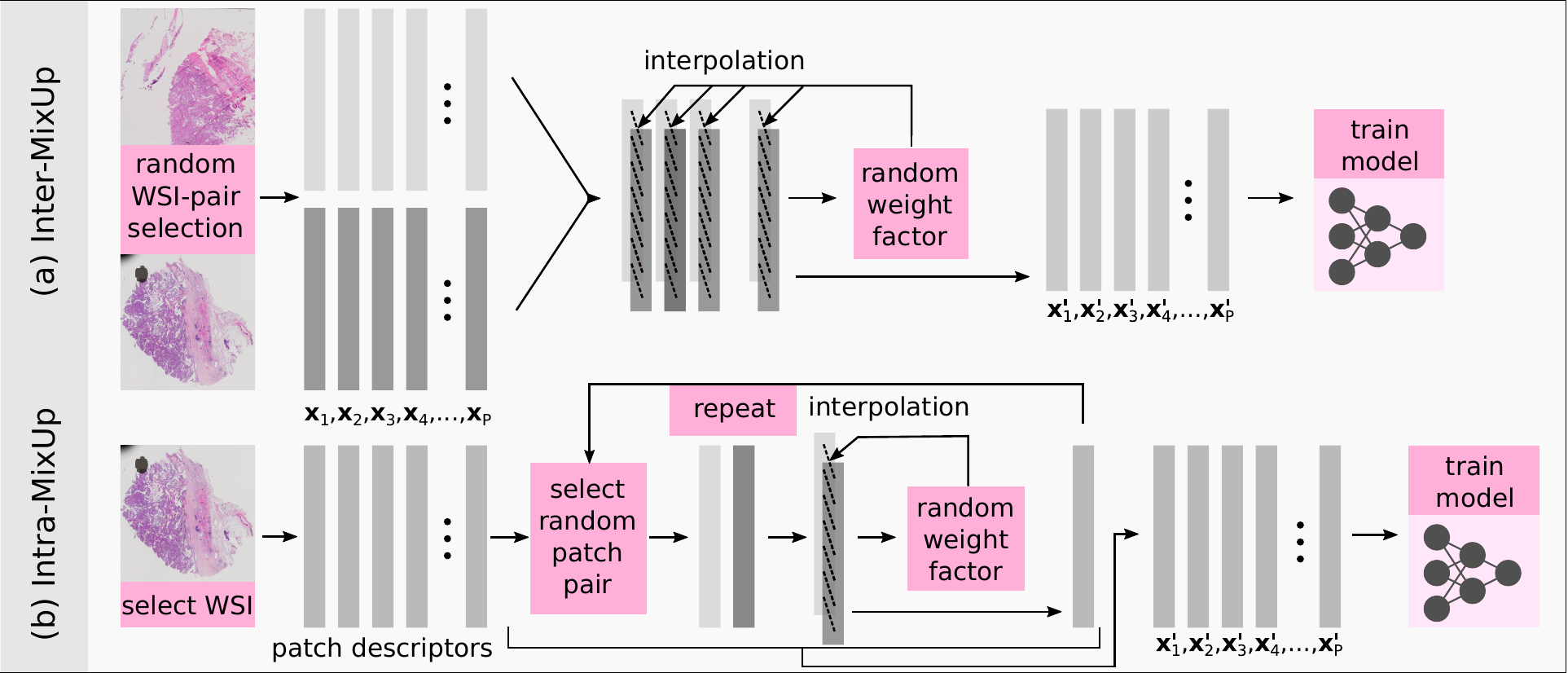}
	\caption{Overview of the proposed feature-based data augmentation approaches. In the case of Inter-MixUp (a), a linear combination was applied on the pairs of WSI descriptors with a randomly selected weight factor. In the case of Intra-MixUp (b), patch-based descriptors from the same WSI were merged with individual random weights.}
	\label{fig:outline}
\end{figure}

\subsection{Inter-MixUp \& Intra-MixUp} 
\textit{Inter-MixUp} refers to the generation of synthetic feature vectors by linearly combining feature vectors of a pair of WSIs (see Fig.~\ref{fig:outline}~(a)).
All features of a WSI with index $w$ can be represented by $X^{(w)}$, such that
${X^{(w)}} = (\boldsymbol{x^{(w)}_{1}}, \; ... \;, \boldsymbol{x^{(w)}_{P}}) \; .$ 
\noindent
To generate a new synthetic sample ${X^{(u)}}'$ based on two samples ${X^{(w)}}$ and ${X^{(v)}}$, we introduce the operation 
$${X^{(u)}}' = 
(\alpha \cdot \boldsymbol{x_1^{(w)}} + (1-\alpha) \cdot \boldsymbol{x^{(v)}_1}, \; \alpha \cdot \boldsymbol{x^{(w)}_2}+ (1-\alpha) \cdot \boldsymbol{x^{(v)}_2}, \; ... \;, \;  \alpha \cdot \boldsymbol{x^{(w)}_P} + (1-\alpha) \cdot \boldsymbol{x^{(v)}_P}) $$
with $\alpha$ being a uniformly sampled random weight ($\alpha \in [0,1]$).
The WSI indexes $v$ and $w$ are uniformly sampled from the set of indexes. The index $u$ ranges from the 1 to the number of extracted WSI descriptors. Since the new synthetic descriptors are individually generated in each epoch, there is no benefit if the number of extracted WSI descriptors is increased. We fix this number to the number of WSIs in the training data set, in order to keep the number of training iterations per epoch consistent.

Two different configurations are considered. Firstly, we investigate the interpolation between WSIs of the same class (V1). Secondly, interpolation between all WSIs is performed, which also includes the interpolation between the labels (V2).
In the case of V2, also the one-hot-encoded label vectors are linearly combined, such that
$\boldsymbol{y^{(u)}}' = \alpha \cdot \boldsymbol{y^{(w)}} + (1-\alpha) \cdot \boldsymbol{y^{(v)}}$ 
The random values, $\alpha$, $v$ and $w$ are selected individually for each individual WSI and each epoch. 
Before applying the MixUp operation, the vector tupel is randomly shuffled (as performed in all experiments).

Intra-WSI combinations (\textit{Intra-MixUp}) refers to the generation of synthetic descriptors by combining feature vectors within an individual WSI (see Fig.~\ref{fig:outline}~(b)).
A new synthetic patch descriptor $\boldsymbol{x_k}'$ is created based on the randomly selected descriptors $\boldsymbol{x_i}$ and $\boldsymbol{x_j}$, such that
$\boldsymbol{x_k}'= \alpha \cdot \boldsymbol{x_i} + (1-\alpha) \cdot \boldsymbol{x_j} \; , $ 
with $i$ and $j$ being random indices (uniformly sampled from $\{1, 2, ..., P\}$) and $\alpha$ being a uniformly sampled random value a ($\alpha \in [0,1]$). The index $k$ ranges from 1 to the number of extracted descriptors per patch. This number was kept stable (1024) during all experiments.
The thereby obtained vector tupel $(\boldsymbol{x_1}', ..., \boldsymbol{x_P}')$ finally represents the synthetic WSI-based image descriptor.
Besides performing combinations for each WSI during training, selective interpolation can be useful to keep real samples within the training data. This can be easily achieved by choosing 
$(\boldsymbol{x_1}', ..., \boldsymbol{x_P}')$ 
with a chance of $\beta$ and $(\boldsymbol{x_1}, ..., \boldsymbol{x_P})$ otherwise. 
While the Intra-MixUp method described before represents a linear interpolation method, we also investigated a multilinear approach by computing $\boldsymbol{x_k}'$ such that 
$\boldsymbol{x_k}'= \boldsymbol{\alpha} \circ \boldsymbol{x_i} + (1-\boldsymbol{\alpha}) \circ \boldsymbol{x_j}$ with $\boldsymbol{\alpha}$ being a random vector and $\circ$ being the element-wise product. This element-wise linear (multilinear) approach enables even higher variability in the generated samples.

\subsection{Experimental Setting}

As experimental architecture, use the dual-stream MIL approach proposed by Li et al~\cite{Li21a}.
Since this model combines both, embedding-based and an instance-based encoding, the effect of both paths can be individually investigated without changing any other architectural details. Since the method represents a state-of-the-art approach, it further serves as well-performing baseline.
In instance-based MIL, the information per patch is first condensed to a single scalar value, representing the classification per patch. Finally, all of these patch-based values are aggregated. 
In embedding-based MIL, the information per patch is translated into a feature vector. All feature vectors from a WSI are then aggregated followed by a classification.
In the investigated model~\cite{Li21a} an instance- and an embedding-based pathway are employed in parallel and are merged in the end by weighted addition. 
The embedding-based pathway contains an attention mechanism
, to higher weight patches that are similar to the so-called critical instance. The model makes use of an individual feature extraction stage.
Due to the limited number of WSIs, we did not train the feature extraction stage~\cite{Hou16a}, but utilize a pre-trained network instead. 
Specifically, we applied a ResNet18 pre-trained on the image-net challenge data, due to the high performance in previous work on similar data~\cite{Gadermayr21a}.
ResNet18 was assessed as particularly appropriate due to the rather low dimensional output (512 dimensions).
We actively decided not to use a self-supervised contrastive learning approach~\cite{Li21a} as feature extraction stage since invariant features could interfere with the effect of data augmentation.
We investigated various settings consisting of instance-based only (INST), embedding-based only (EMB) and the dual-stream approach with weightings 3/1, 2/2 (balanced) and 1/3 for the instance and the embedding-based pathways.

As comparison, several other augmentation methods on feature level are investigated including random sampling, selective random sampling and random noise. 
%
Random sampling corresponds to the random selection of patches (feature vectors) from each WSI. Thereby the amount of investigated data per WSI is reduced with the benefit of increasing the variability of the data. In the experiments, we adjust the sample ratio $q$ between the patch-based features for training and testing. A $q$ of 50 \% indicates that 512 descriptors are used for training while for testing always a fixed number of 1024 is used.
%
Selective random sampling corresponds to the random sampling strategy, with the difference that the ratio of features is not fixed but drawn from a uniform random distribution ($U(q,100 \; \%)$). Here, a $q$ of $50 \; \%$ indicates that for each WSI, between $512$ and $1024$ feature vectors are selected.
%
In the case of the random noise setting, to each feature vector $\boldsymbol{x_i}$, a random noise vector $\boldsymbol{r}$ is added
($\boldsymbol{x_i}' = \boldsymbol{x_i} + \boldsymbol{r} $). The elements of $r$ are randomly sampled (individually for each $\boldsymbol{x_i}$) from a normal distribution $N(0,\sigma')$. To incorporate for the fact that the feature dimensions show different magnitudes, $\sigma'$ is computed as the product of the meta parameter $\sigma$ and the standard deviation of the respective feature dimension.

In this work, we aimed at distinguishing different nodular lesions of the thyroid, focusing especially on benign follicular nodules (FN) and papillary carcinomas (PC).
This differentiation is crucial, due to the different treatment options, in particular with respect to the extent of surgical resection of the thyroid gland~\cite{Xi2022}.
The data set utilized in the experiments consists of 80 WSIs overall. One half (40) of the data set consists of frozen and the other half (40) of paraffin sections~\cite{Gadermayr21a}), representing the different modalities. All images were acquired during clinical routine at the 
Kardinal Schwarzenberg 
Hospital. Procedures were approved by the ethics committee of the county of Salzburg (No. 1088/2021). The mean and median age of patients at the date of dissection was 47 and 50 years, respectively. The data set comprised 13 male and 27 female patients, corresponding to a slight gender imbalance.
They were labeled by an expert pathologist with over 20 years experience. A total of 42 (21 per modality) slides were labeled as papillary carcinoma while 38 (19 per modality) were labeled as benign follicular nodule.
For the frozen sections, fresh tissue was frozen at $-15^\circ$ Celsius, slides were cut (thickness 5 $\mu m$) and stained immediately with hematoxylin and eosin.
For the paraffin sections, tissue was fixed in $4 \; \%$ phosphate-buffered formalin for 24 hours. Subsequently formalin fixed paraffin embedded tissue was cut (thickness 2 $\mu m$) and stained with hematoxylin and eosin.
The images were digitized with an Olympus VS120-LD100 slide loader system. Overviews at a 2x magnification were generated to manually define scan areas, focus points were automatically defined and adapted if needed. Scans were performed with a 20x objective (corresponding to a resolution of 344.57 nm/pixel). The image files were stored in the Oympus vsi format based on lossless compression.

The data set was randomly separated into training ($80 \; \%$) and test data ($20 \; \%$). 
The whole pipeline, including the separation, was repeated 32 times to achieve representative scores. 
Due to the almost balanced setting, the overall classification accuracy (mean and standard deviation) is finally reported.
Adam was used as optimizer. The models were trained for 200 epochs with an initial learning rate of 0.0002.
Random shuffling of the vector tupels (shuffling within the WSIs) was applied for all experiments.
\begin{figure}
	\includegraphics[width=\linewidth]{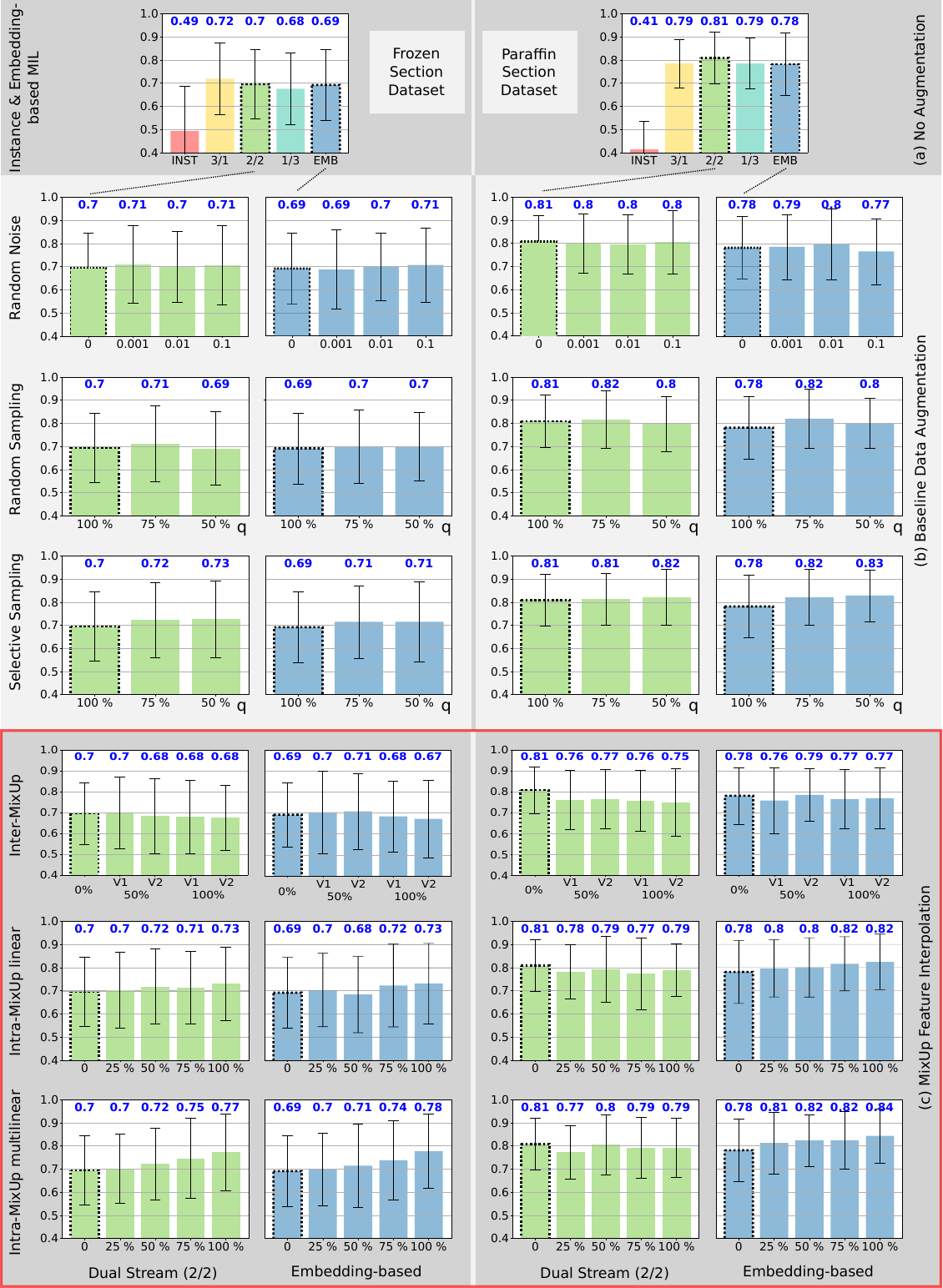}
	\caption{Mean overall classification accuracy and standard deviation obtained with each individual combination. The columns represent the frozen (left) and paraffin data set (right). The top row (a) shows the baseline scores of embedding-based, instance-based and 3 combinations. Subfigure (b) shows the scores obtained with baseline data augmentation for embedding-based and dual-stream MIL. Subfigure (c) shows the scores obtained with interpolation between (Inter-MixUp) and within WSIs (Intra-MixUp).}
	\label{fig:res}
\end{figure}
The patches were randomly extracted from the WSI, based on uniform sampling. For each patch, we checked that at least $75 \; \%$ of the area was covered with tissue (green color channel) in order to exclude empty areas~\cite{Gadermayr21a}. To obtain a representation independent of the WSI size, we extracted 1024 patches with a size of $256 \times 256$ pixel per WSI, resulting in 1024 patch-descriptors per WSI~\cite{Gadermayr21a}.
For feature extraction, a ResNet18 network, pretrained on the image-net challenge was deployed~\cite{Li21a}. Data and source code are publicly accessible via 
\href{https://gitlab.com/mgadermayr/mixupmil}{\url{https://gitlab.com/mgadermayr/mixupmil}}. 
We use the reference implementation of the dual-stream MIL approach~\cite{Li21a}.
To obtain further insight into the feature distribution, we randomly selected patch descriptor pairs and computed the Euclidean distances. In detail, we selected 10,000 pairs 
(a) from different classes, 
(b) from different WSIs (similar and dissimilar classes), 
(c,d) from the same class and different WSIs, and
(e) from the same WSI.

\section{Results} \label{sec:results}

\begin{figure}[tb]
    \centering
    \includegraphics[width=0.99\linewidth]{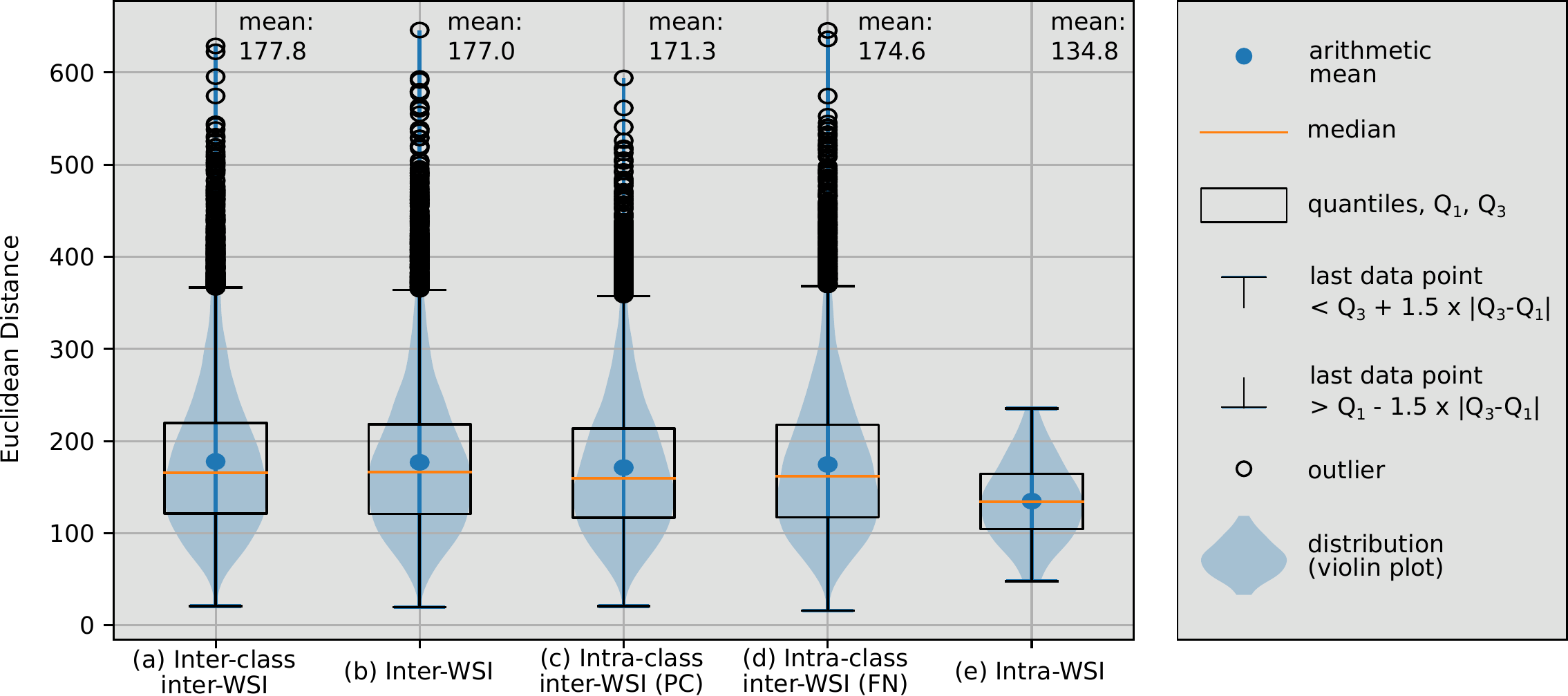}
    \caption{Analysis of the distributions of the patch descriptor distances between (a) patches from different classes, (b) randomly selected patches from different WSIs, (c,d) patches from the same class and different WSIs (for both classes, PC and FN) and (e) patches within the WSIs.}
    \label{fig:boxviolin}
\end{figure}

Figure~\ref{fig:res} shows the mean overall classification accuracy and standard deviations obtained with each individual combination. The columns represent the frozen (left) and paraffin data set (right). The top row (a) shows the baseline scores of embedding-based, instance-based and the 3 combinations. Subfigure (b) show the scores obtained with baseline data augmentation for embedding-based and dual-stream MIL. Subfigure (c) shows the scores obtained with interpolation between patches between (Inter-MixUp) and within WSIs (Intra-MixUp).
Without data augmentation, scores between 0.49 and 0.72 were obtained for frozen and scores between 0.41 and 0.81 for the paraffin data set.
To limit the number of figures and due to the fact that instance-based MIL showed weak scores only, in the following part the focus is on embedding-based and combined-MIL (2/2) only.
With baseline data augmentation, scores between 0.69 and 0.73 were achieved for the frozen and between 0.78 and 0.83 for the paraffin data set.
Inter-MixUp exhibited scores up to 0.71 for the frozen and up to 0.79 for the paraffin data set.
Intra-MixUp showed average accuracy up to 0.78 for the frozen and up to 0.84 for the paraffin data set. The best scores were obtained with the multilinear setting.
In Fig.~\ref{fig:boxviolin}, the distributions of the descriptor (Euclidean) distances between (a-d) patches from different different WSIs (inter-WSI) and (e) patches within a single WSI (intra-WSI) are provided. The mean distances range from 171.3 to 177.8 for the inter-WSI settings. In the intra-WSI setting, a mean distance of 134.8 was obtained.
Based on the used common box plot variation (whiskers length is less than $1.5 \times$ the interquartile range), a large number of data points was identified as outliers. However, these points are not considered as real outliers, but occur due to the asymmetrical data distribution (as indicated by the violin plot in the background).

\section{Discussion}

In this work, we proposed and examined novel data augmentation strategies based on the idea of interpolations of feature vectors in the MIL setting.
Instance-based MIL did not show any competitive scores. Obviously the model reducing each patch to a single value is not adequate for the classification of frozen or paraffin sections from thyroid cancer tissues.
The considered dual-stream approach, including an embedding and instance-based stream, exhibited slightly improved average scores, compared to embedding-based MIL only.
In our analysis, we focused on the embedding-based configuration and on the balanced combined approach (referred to as 2/2).
With the baseline data augmentation approaches, the maximum improvements were 0.03, and 0.02 for the frozen, and 0.01, and 0.05 for the paraffin data set.
The Inter-MixUp approach did not show any systematic improvements. Independently of the chosen strategy (V1, V2), concerning the combination within or between classes, we did not notice any positive trend. 
The multilinear Intra-MixUp method, however, exhibited the best scores for 3 out of 4 combinations and the best overall mean accuracy for both, the frozen and the paraffin data set. Also a clear trend with increasing scores in the case of an increasing ratio of augmented data ($\beta$) is visible. The linear method showed a similar, but less pronounced trend.
Obviously, the straightforward application of the MixUp scheme (as in case of the Inter-MixUp approach), is inappropriate for the considered setting. An inhibiting factor could be a high inter-WSI variability leading to incompatible feature vectors (which are too far away from realistic samples in the feature space). To particularly investigate this effect, we performed 2 different Inter-MixUp settings (V1 \& V2), with the goal of identifying the effect of mixed (and thereby more dissimilar) or similar classes during interpolation. 
%
The analysis of the distance distributions between patch representations confirmed that, the variability between WSIs is clearly larger than the variability within WSIs. In addition, the results showed that the variability between classes is, on patch-level, not clearly larger than the variability within a class.
Obviously variability due to the acquisition outweigh any disease specific variability.
This could provide an explanation for the effectiveness of Intra-MixUp approach compared to the (similarly) poorly performing Inter-MixUp settings.
We expect that stain normalization methods (but not stain augmentation) could be utilized to align the different WSIs to provide a more appropriate basis for inter-WSI interpolation.
%
%
%
With regard to the different data sets, we noticed a stronger, positive effect in case of the frozen section data set. This is supposed to be due to the clearly higher variability of the frozen sections corresponding with a need for a higher variability in the training data. 
We also noticed a stronger effect of the solely embedding-based architecture (also showing the best overall scores). We suppose that this is due to the fact that the additional loss of the dual-stream architecture exhibits a valuable regularization tool to reduce the amount of needed training data. With the proposed Intra-MixUp augmentation strategy, this effect diminishes, since the amount and quality of training data is increased.


To conclude, we proposed novel data augmentation strategies based on the idea of interpolations of image descriptors in the MIL setting.
Based on the experimental results, the multilinear Intra-MixUp setting proved to be highly effective, while the Inter-MixUp method showed inferior scores compared to a state-of-the-art baseline.
We learned that there is a clear difference between combinations within and between WSIs with a noticeable effect on the final classification accuracy.
This is supposedly due to the high variability between the WSIs compared to a rather low variability within the WSIs.
In the future, additional experiments will be conducted including stain normalization methods and larger benchmark data sets to provide further insights. 


\subsection*{Acknowledgement}
This work was partially funded by the County of Salzburg (no. FHS2019-10-KIAMed)


\bibliographystyle{splncs04}

\bibliography{bib.bib}

\end{document}